\documentclass[twoside]{article}

\usepackage[accepted]{aistats2020}

\usepackage[utf8]{inputenc} \usepackage[T1]{fontenc}    \usepackage{hyperref}       \usepackage{url}            \usepackage{booktabs}       \usepackage{amsfonts}       \usepackage{nicefrac}       \usepackage{microtype}      

\usepackage{booktabs} \usepackage{amsmath}
\usepackage{amssymb}
\usepackage{amsthm}
\usepackage{tikz}
\usetikzlibrary{arrows,patterns}
\usepackage{bm}
\usepackage{pgfplots,xcolor}
\pgfplotsset{compat=1.12} 
\usepgfplotslibrary{ternary}
\usepackage{cleveref}
\usepackage{wrapfig}
\usepackage{subcaption}
\usepackage[]{natbib}
\usepackage{enumerate}
\usepackage{needspace}

\newtheorem{prop}{Proposition}
\newtheorem{defn}{Definition}

\newcommand{\Dalpha}[2]{\ensuremath{D_{\alpha}(#1 \,\|\, #2)}}
\newcommand{\hatDalpha}[2]{\ensuremath{\hat{D}_{\alpha}(#1 \,\|\, #2)}}
\newcommand{\Dkl}[2]{\ensuremath{D_{\textrm{KL}}(#1 \,\|\, #2)}}
\newcommand{\Dinfty}[2]{\ensuremath{D_{\infty}(#1 \,\|\, #2)}}

\newcommand{\Pd}{{{P}}}
\newcommand{\Qd}{{{Q}}}
\newcommand{\Rd}{{{R}}}
\newcommand{\bmu}{{\bm{\mu}}}

\newcommand{\bp}{{\mathbf{p}}}
\newcommand{\bq}{{\mathbf{q}}}
\newcommand{\br}{{\mathbf{r}}}

\newcommand{\pd}{{\mathbf{p}}}
\newcommand{\qd}{{\mathbf{q}}}
\newcommand{\rd}{{\mathbf{r}}}

\newcommand{\supp}{{\textrm{supp}}}
\newcommand{\btheta}{{\bm{\theta}}}
\newcommand{\R}{{\mathbb{R}}}
\DeclareMathOperator*{\argmin}{arg\,min}
\newcommand{\PRD}{{\textsc{PRD}}}

\newcommand{\Ffwd}{{\ensuremath{{\mathcal{F}^\cap}}}}
\newcommand{\Frev}{{\ensuremath{{\mathcal{F}^\cup}}}}

\newcommand{\myparagraph}[1]{\textbf{#1}\quad}

\hypersetup{
    colorlinks,
    anchorcolor={blue!90!black},
    linkcolor={blue!90!black},
    citecolor={blue!90!black},
    urlcolor={blue!90!black}
}

\newenvironment{sketch}{\proof}{\endproof}

\begin{document}

\twocolumn[

\aistatstitle{Precision-Recall Curves Using Information Divergence Frontiers}

\aistatsauthor{ Josip Djolonga \And Mario Lucic \And  Marco Cuturi \AND Olivier Bachem \And Olivier Bousquet \And Sylvain Gelly}

\aistatsaddress{ Google Research, Brain Team} ]

\begin{abstract}
Despite the tremendous progress in the estimation of generative models, the development of tools for diagnosing their failures and assessing their performance has advanced at a much slower pace. Recent developments have investigated metrics that quantify which parts of the true distribution is modeled well, and, on the contrary, what the model fails to capture, akin to precision and recall in information retrieval. In this paper, we present a general evaluation framework for generative models that measures the trade-off between precision and recall using R\'enyi divergences.
Our framework provides a novel perspective on existing techniques and extends them to more general domains.
As a key advantage, this formulation encompasses both continuous and discrete models and allows for the design of efficient algorithms that do not have to quantize the data.
We further analyze the biases of the approximations used in practice.
\end{abstract}

\vspace{-3mm}
\section{INTRODUCTION}

Deep generative models, such as generative adversarial networks \citep{goodfellow2014generative} and variational autoencoders \citep{kingma2013auto,rezende2014stochastic}, have recently risen to prominence due to their ability to model high-dimensional complex distributions. While we have witnessed a tremendous growth in the number of proposed models and their applications, a comprehensive set of quantitative evaluation measures is yet to be established. Obtaining sample-based quantities that can reflect common issues occurring in generative models, such as ``mode dropping'' (failing to adequately capture all the modes of the target distribution) or ``oversmoothing'' (inability to produce the high frequency characteristics of points in the true distribution) remains a key research challenge.

Currently used metrics, such as the inception score (IS) \citep{salimans2016improved} and the Fr\'echet inception distance (FID) \cite{heusel2017gans} produce single number summaries quantifying the goodness of fit.
Thus, even though they can detect poor performance, they cannot shed light upon the underlying cause.
\citet{sajjadi2018assessing} and later \citet{kynkaanniemi2019improved} have offered an alternative view, motivated by the notions of precision and recall in information retrieval. Intuitively,  the precision captures the average ``quality'' of the generated samples, while the recall measures how well the target distribution is covered.
They have demonstrated that such metrics can disentangle these two common failure modes on a set of image synthesis experiments.

Unfortunately, these recent approaches rely on data quantization and do not provide a theory that can be directly used on with continuous distributions.
For example, in \cite{sajjadi2018assessing} the data is first clustered and then the resulting class-assignment histograms are compared.
Recently, \citet{simon2019revisiting} suggest an algorithm that extends to the continuous setting by using a density ratio estimator, a result that we extend to arbitrary R\'enyi divergences.
In \cite{kynkaanniemi2019improved} the space is covered with hyperspheres and is only sensitive to the size of the overlap of the supports of the distributions.

In this work, we present an evaluation framework based on the Pareto frontiers of R\'enyi divergences that encompasses these previous contributions as special cases. Beyond this novel perspective on existing techniques, we provide a general characterization of these Pareto frontiers, in both the discrete and continuous case. This in turn enables efficient algorithms that are directly applicable to continuous distributions without the need for discretization.

\myparagraph{Contributions} (1) We propose a general framework for comparing distributions based on the Pareto frontiers of statistical divergences. (2) We show that the family of R\'enyi divergences are particularly well suited for this task and produce curves that can be interpreted as precision-recall trade-offs.
(3) We develop tools to compute these curves for several widely used families of distributions.
(4) We show that the recently popularized definitions of precision and recall \citep{sajjadi2018assessing,kynkaanniemi2019improved} correspond to specific instances of our framework. In particular, we give a theoretically sound geometric interpretation of the definitions and algorithms in \citep{sajjadi2018assessing,kynkaanniemi2019improved}.
(5) We analyze the consequences of the approximations made when these methods are used in practice.

\begin{figure*}[t]
    \centering
    \begin{subfigure}[t]{0.55\textwidth}
    \centering
    \includegraphics[width=0.7\textwidth]{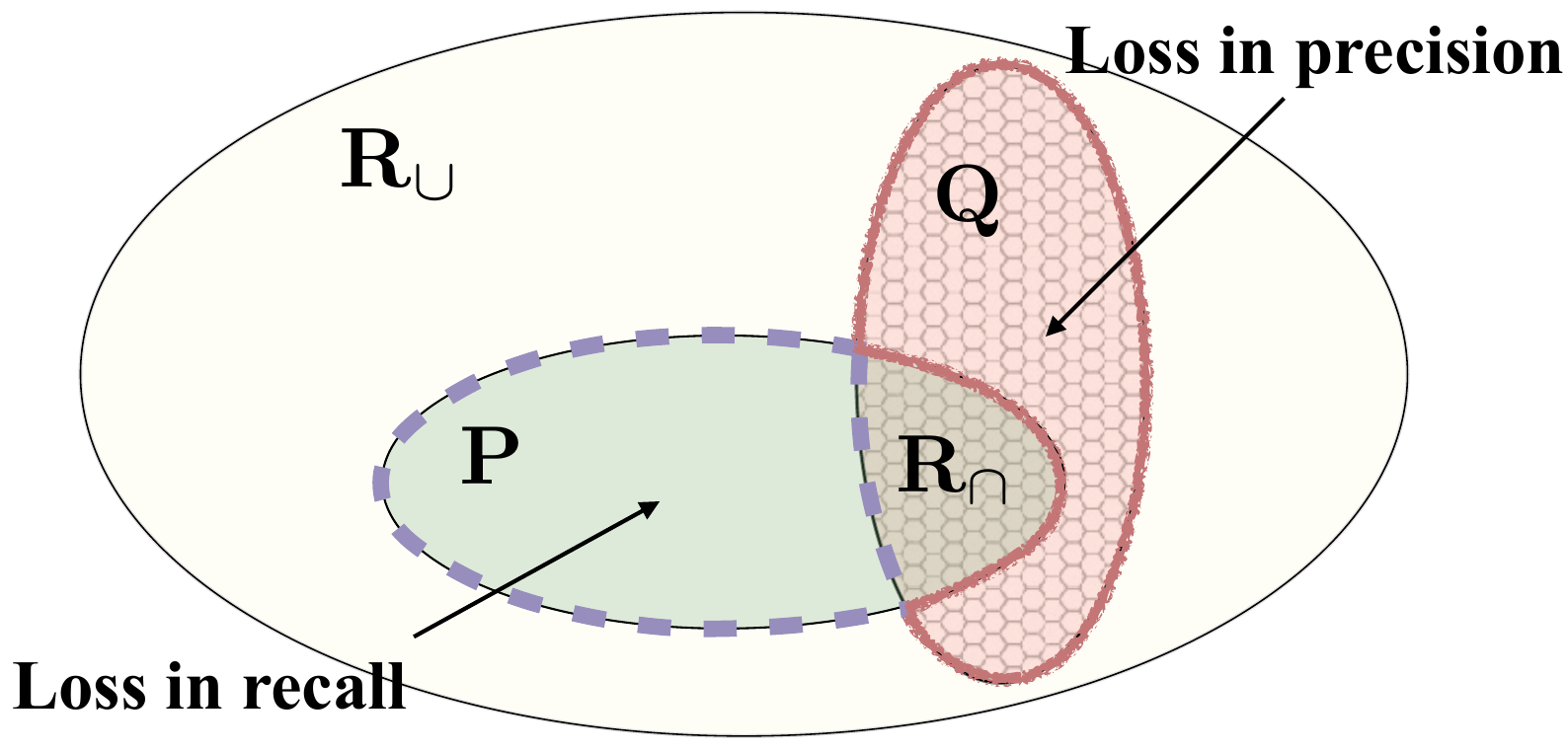}
    \caption{When $\Pd$ and $\Qd$ are uniform we can define natural precision and recall concepts using $\Rd_\cup$ and $\Rd_\cap$, which are uniform on the union and intersection of the supports of $\Pd$ and $\Qd$ respectively.}
    \label{fig:intuition_uniform}
    \end{subfigure}\hspace*{.3cm}
    \begin{subfigure}[t]{0.4\textwidth}
    \centering
    \includegraphics[width=0.8\textwidth]{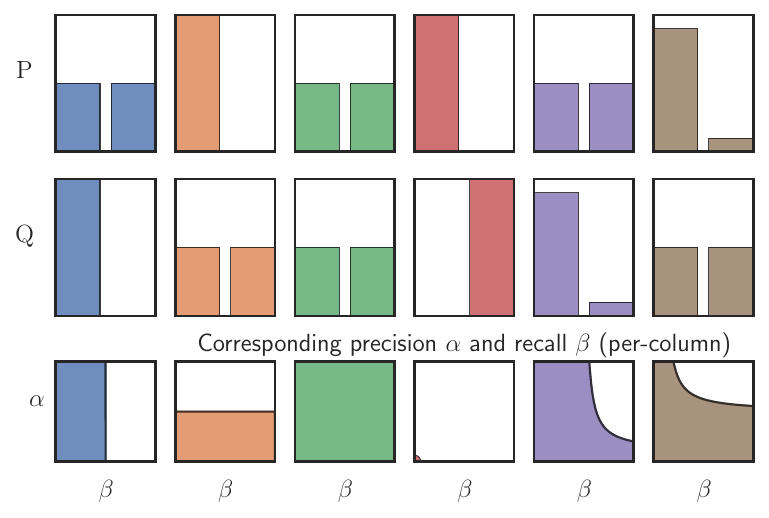}
    \subcaption{Examples with categorical $\Pd$ and $\Qd$, reproduced from \cite{sajjadi2018assessing}.}
    \label{fig:intuition_categorical}
        \end{subfigure}
  \caption{For uniform measures we can define natural concepts using set operation (a). Similarly, when they are simple categorical distributions, we would like to generate curves like those in (b). }
  
  \label{fig:intuition}
\end{figure*}

The central problem considered in the paper is the development of a framework that formalizes the concepts of precision and recall for arbitrary measures, and enables the development of principled evaluation tools.
Namely, we want to understand how does a learned model, henceforth denoted by $\Qd$, compare to the target distribution $\Pd$.
Informally, to compute the precision we need to estimate how much probability $\Qd$ assigns to regions of the space where $\Pd$ has high probability. Alternatively, to compute the recall we need to estimate how much probability $\Pd$ assigns to regions of the space that are likely under $\Qd$.

Let us start by developing an intuitive understanding of the problem with simple examples where the relationship between $\Pd$ and $\Qd$ is easily understandable.
\Cref{fig:intuition} illustrates the case where $\Pd$ and $\Qd$ are uniform distributions with supports $\supp(\Pd)$ and $\supp(\Qd)$.
To help with the exposition of our approach in the next section, we also introduce the distributions $\Rd_\cup$ and $\Rd_\cap$ which are uniform on the union and intersection of the supports of $\Pd$ and $\Qd$ respectively.
Then, the \emph{loss in precision} can then be understood to be proportional to the measure of $\supp(\Qd) \setminus \supp(\Rd_\cap)$ which corresponds to the "part of $\Qd$ not covered by $\Pd$".
Analogously, the \emph{loss in recall} of $\Qd$ w.r.t.\ $\Pd$ is proportional to the size of $\supp(\Pd) \setminus \supp(\Rd_\cap)$ which represents the "part of $\Pd$ not covered by $\Qd$.
Note that we can also write these sets as $\supp(\Rd_\cup)\setminus \supp(\Pd)$ and $\supp(\Rd_\cup)\setminus \supp(\Qd)$ respectively.
The precision and recall are then naturally maximized when $\Pd=\Qd$.
When the distributions are discrete we would like to generate plots similar to those in  \Cref{fig:intuition_categorical}. The first column corresponds to $\Qd$ which fails to model one of the modes of $\Pd$, and the second column to a $\Qd$ which has an ``extra'' mode.
We would like our framework to mark these two failure modes as losses in recall and precision, respectively. The third column corresponds to $\Pd=\Qd$, followed by a situation where $\Pd$ and $\Qd$ have disjoint support. Finally, for the last two columns, a possible precision-recall trade-off is illustrated. While this intuition is satisfying for uniform and categorical distributions, it is unclear how to extend it to continuous distributions that might be supported on the complete space. 

 \section{DIVERGENCE FRONTIERS}

To formally transport these ideas to the general case, we will introduce an auxiliary distribution $\Rd$ that is constrained to be supported only on those regions where both $\Pd$ and $\Qd$ assign high probability\footnote{This is in contrast to \citep{sajjadi2018assessing}, who require $\Pd$ and $\Qd$ to be mixtures with a shared component.}.
Informally, this should act as a generalization to the general case of $\Rd_\cap$, which was the measure on the intersection of the supports of $\Pd$ and $\Qd$.
Then, the \emph{discrepancy} between $\Pd$ and $\Rd$ measures the space that is likely under $\Pd$ but not under $\Rd$, which can be seen as loss in recall. Similarly, the discrepancy between $\Qd$ and $\Rd$ quantifies the size of the space where $\Qd$ assigns probability mass, but $\Pd$ does not, which we can be interpreted as a loss in recall.

Hence, we need both a mechanism to measure distances between distributions and means to constrain $\Rd$ to assign mass only where both $\Pd$ and $\Qd$ do. For example, if $\Pd$ and $\Qd$ are both mixtures of several components $\Rd$ should assign mass only to the components shared by both $\Pd$ and $\Qd$.

\myparagraph{A dual view}
Alternatively, building on the observation from the previous section that both $\Rd_\cup$ and $\Rd_\cap$ can be used to define precision and recall, instead of modeling the intersection of $\Pd$ and $\Qd$, we can use an auxiliary distribution $\Rd$ to approximate the \emph{union} of the high-probability regions of $\Pd$ and $\Qd$. Then, using a similar analogy as before, the distance between $\Pd$ and $\Rd$ should measure the loss in precision, while the distance between $\Qd$ and $\Rd$ the loss in recall.
In this case, $\Rd$ should give non-zero probability to any part of the space where either $\Pd$ or $\Qd$ assign mass.
When $\Pd$ and $\Qd$ are both mixtures of several components, $\Rd$ has to be supported on the union of all mixture components.

As a result, the choice of the statistical divergence between $\Pd$, $\Qd$ and $\Rd$ becomes paramount.

\subsection{Choice of Divergence}
To be able to constrain $\Rd$ to assign probability mass only in those regions where $\Pd$ and $\Qd$ do, we need a measure of discrepancy between distributions that penalizes differently under- and over-coverage.
Even though the theory and concepts introduced in this paper extend to any such divergence, we will focus our attention to the family of R\'enyi divergences.
They not only do exhibit such behavior, but their properties are also well-studied in the literature, which we can leverage to develop a deeper understanding of our approach, and in the design of efficient computational tools.

\begin{figure}[ht] \centering
  \includegraphics[width=.23\textwidth]{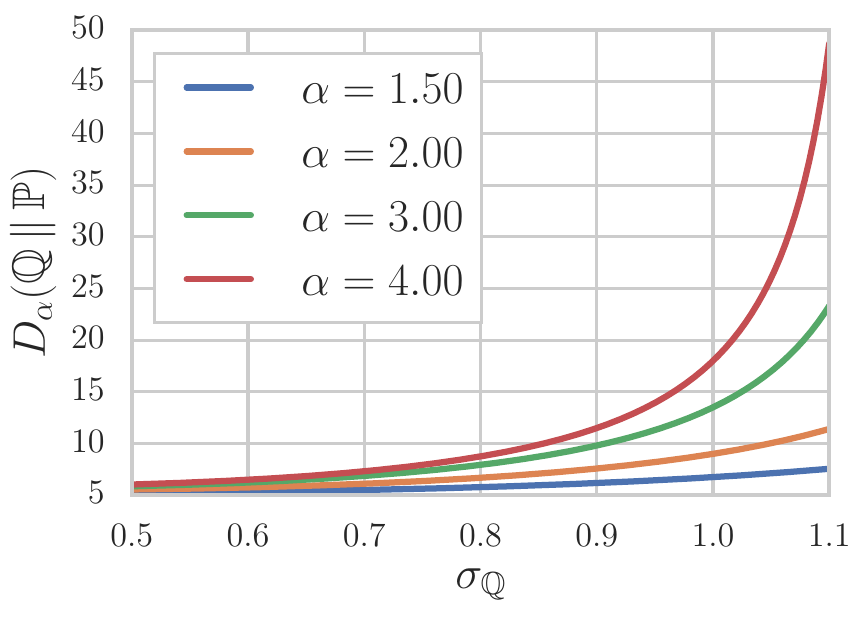}
  \includegraphics[width=.23\textwidth]{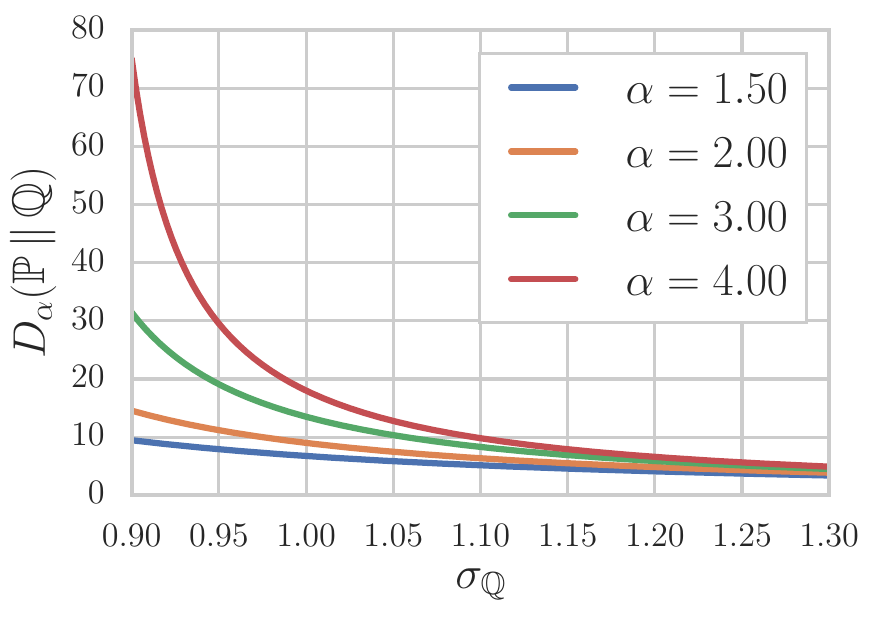}
  \caption{R\'enyi divergences strongly penalize when the first argument assigns mass away from the high probability regions of the second. We analytically evaluate $D_\alpha$, where $\mathbb{P}$ is a Normal \citep{gil2013renyi}.\label{fig:inclusive_exclusive}}
\end{figure}

\begin{defn}[R\'enyi Divergence \citep{renyi}] Let $\Pd$ and $\Qd$ be two measures such
  that $\Qd$ is absolutely continuous with respect to $\Pd$, i.e., any measure set with zero measure under $\Pd$ has also
  zero measure under $\Qd$. Then, the R\'enyi divergence of order
  $\alpha\in(0,1)\cup(1,\infty)$ is defined as
  \begin{equation} \Dalpha{\Pd}{\Qd} = \frac{1}{\alpha-1} \log \int
    \left(\frac{d\Pd}{d\Qd}\right)^{\alpha-1} d\Pd,
  \end{equation} where $d\Pd/d\Qd$ is the Radon-Nikodym derivative\footnote{Equal to the ratios of the densities of $\Pd$ and $\Qd$ when they both exist.}.
\end{defn}
The fact that they are sensitive to how the supports of $\Pd$ and $\Qd$ relate to one another is already hinted by the constraint in the definition, which requires that $\supp(\Pd)\subseteq\supp(\Qd)$.
Furthermore, by increasing the parameter $\alpha$ the divergence becomes ``less forgiving'' --- for example if $\Pd$ and $\Qd$ are Gaussians with deviations $\sigma_\Pd$ and $\sigma_\Qd$, we have that $\Dalpha{\Pd}{\Qd}$ increases faster as $\alpha\to\infty$ when $\sigma_\Qd$ drops below $\sigma_\Pd$, while $\Dalpha{\Qd}{\Pd}$ grows with increasing $\sigma_\Qd$ and $\alpha\to\infty$, which we illustrate in \Cref{fig:inclusive_exclusive}.
This is exactly the property that we need to be able to  define meaningful concepts of precision and recall.
For a more detailed analysis of this behavior we point the reader to \cite{minka2005divergence}.

R\'enyi divergences have been extensively studied in the literature~\citep{erven2014renyi} and many of their properties are well-understood --- for example, they are non-negative and zero only if the distributions are equal a.s., and increasing in $\alpha$.  Some of their orders are closely related to the Hellinger and $\chi^2$ divergences, and it can be further shown that $\Dkl{\Pd}{\Qd}=\int \log(\frac{d\Pd}{d\Qd})d\Pd = \lim_{\alpha\to 1}\Dalpha{\Pd}{\Qd}$.

\subsection{Divergence Frontiers}

Having defined a suitable discrepancy measure, we are ready to define the central concepts in this paper, which will play the role of precision-recall curves for arbitrary measures.
To do so, we will not put hard constraints on $\Rd$, but only softly enforce them.
Namely, consider the case when we want $\Rd$ to model the intersection of the high likelihood regions of $\Pd$ and $\Qd$.
Then, if it fails to do so, either $\Dalpha{\Rd}{\Pd}$ or $\Dalpha{\Rd}{\Qd}$ will be significantly large.
Similarly, unless $\Rd$ fails to assign large probabilities to the high likelihood regions of both $\Pd$ and $\Qd$, at least one of $\Dalpha{\Pd}{\Rd}$ and $\Dalpha{\Qd}{\Rd}$ will be large.
Thus, we  will only consider those $\Rd$ that simultaneously minimize both divergences, which motivates the following definition.

\begin{defn}[Divergence frontiers]\label{defn:main} For any two measures $\Pd$ and $\Qd$, any class of measures $\mathcal{M}$ and any
  $\alpha\geq 0$, we define the exclusive realizable region as the set
  \begin{equation}\label{eqn:region}
    {{\mathcal{R}}}^\cap_\alpha(\Pd, \Qd) = \{ (\Dalpha{\Rd}{\Qd}, \Dalpha{\Rd}{\Pd}) \mid \Rd\in\mathcal{M} \},
  \end{equation} and the inclusive realizable region
  $\mathcal{R}^\cup_\alpha(\Pd, \Qd)$ by swapping the
  arguments of $D_\alpha$ in \eqref{eqn:region}. The exclusive and inclusive divergence frontiers
  are then defined as the maximal points of the corresponding realizable regions\begin{align*}
  {\mathcal{F}}^{\cap}_\alpha(\Pd, \Qd \mid \mathcal{M}) = &\{
    (\pi, \rho) \in
    {\mathcal{R}}^{\cap}_\alpha(\Pd, \Qd \mid \mathcal{M})
    \\ &\mid \nexists
    (\pi',\rho')\in {\mathcal{R}}^{\cap}(\Pd, \Qd)
    \\
    & \textrm{ s.t.\ } \pi' < \pi \textrm{ and } \rho'<\rho \},
 \end{align*}
  and $\mathcal{F}^\cup_\alpha$ is defined by replacing $\mathcal{R}^\cap$ with $\mathcal{R}^\cup$.
\end{defn}

In other words, we want to compute the Pareto frontiers of the multi-objective optimization problem with the divergence minimization objectives
$f_1(\Rd)=\Dalpha{\Rd}{\Qd},f_2(\Rd)=\Dalpha{\Rd}{\Pd}$ and
$f_3(\Rd)=\Dalpha{\Qd}{\Rd},f_4(\Rd)=\Dalpha{\Pd}{\Rd}$ respectively.
In machine learning such divergence minimization problems appear in approximate inference.
Interestingly, $f_1$ and $f_2$ are the central object one minimizes in variational inference (VI) \citep[\S 5]{wainwright2008graphical}\citep{li2016renyi}, while $f_3$ and $f_4$ are exactly the objectives in expectation propagation (EP) \citep{minka2001expectation,minka2005divergence}.
Hence, the problem of computing the frontiers can be seen as that of performing VI or EP with two target distributions instead of one.

 \section{CHARACTERIZATION OF THE FRONTIERS}
Having defined the frontiers, we now characterize them, so that we can discuss their computation in the next section.
Remember that to compute the frontiers we have to characterize the subset of $\R^2$ consisting of all pairs $(\Dalpha{\Rd}{\Pd}, \Dalpha{\Rd}{\Qd})$ and $(\Dalpha{\Pd}{\Rd}, \Dalpha{\Qd}{\Rd})$ which are not strictly dominated.
To solve these two multi-objective optimization problem we scalarize them by optimizing the problems
\begin{align}
  \Rd^\cup_{\alpha,\lambda} &= \argmin_{\Rd} \lambda \hatDalpha{\Qd}{\Rd} + (1-\lambda)\hatDalpha{\Pd}{\Rd} \label{eqn:left-barycenter} \\
  \Rd^\cap_{\alpha,\lambda} &= \argmin_{\Rd} \lambda \hatDalpha{\Rd}{\Qd} + (1-\lambda)\hatDalpha{\Rd}{\Pd} \label{eqn:right-barycenter}.
\end{align}
where $\hat{D}_\alpha=\frac{1}{\alpha-1}e^{\frac{D_\alpha}{\alpha-1}}$ is a monotone function of the R\'enyi divergence.
We then vary $\lambda\in[0, 1]$ and plug $\Rd_{\alpha,\lambda}$ back in $D_\alpha$ to obtain the frontier.

Fortunately, this problem can be analytically solved.
The discrete case has been solved by \citet[III]{nielsen2009dual}, and we modify their argument to the continuous case.
\begin{prop}\label{prob:exact-solutions}
    Let $\Pd$ and $\Qd$ be two measures with densities $p$ and $q$ respectively. Then, the distribution minimizing $\eqref{eqn:left-barycenter}$ has density
    \begin{flalign*}
        &\quad r_{\alpha,\lambda}(x) \propto (\lambda q(x)^{1-\alpha} + (1-\lambda) p(x)^{1-\alpha})^{1/(1-\alpha)}.&
    \end{flalign*}
    Similarly, the optimizer of $\eqref{eqn:right-barycenter}$ is minimized at the distribution with density
    \begin{flalign*}
        &\quad r_{\alpha,\lambda}(x) \propto (\lambda q(x)^{\alpha} + (1-\lambda) p(x)^{\alpha})^{1/\alpha}.&
    \end{flalign*}
\end{prop}
\begin{sketch}
In the inclusive case, \eqref{eqn:left-barycenter} is equal to
\[
\hatDalpha{\Rd_{\alpha,\lambda}^\cap}{\Rd} \int (\lambda q(x)^\alpha + (1-\lambda p(x)^\alpha))^{1/\alpha}dx,
\]
which is minimized when $\Rd=\Rd_{\alpha,\lambda}^\cap$ as the first term is a divergence, and the second term is constant with respect to $\Rd$. The exclusive case is analogous.
\end{sketch}

Even thought not the case for general problems, linear scalarization does yield the correct frontier due to the properties of the R\'enyi divergences.

\begin{prop}\label{prop:discrete-frontiers} For any measures $\Pd$ and $\Qd$ with densities $p$ and $q$ respectively, we can compute the exclusive frontier as
    \begin{equation*}
      \Ffwd_\alpha(P, Q) = \{ (\Dalpha{R^\cap_{\alpha,\lambda}}{\Pd}, \Dalpha{R^\cap_{\alpha,\lambda}}{\Qd}) \mid \lambda\in[0, 1] \},
    \end{equation*}
and the inclusive frontier is given as
 \begin{equation*}
      \Frev_\alpha(P, Q) = \{ (\Dalpha{\Pd}{R^\cup_{\alpha,\lambda}}, \Dalpha{\Qd}{R^\cup_{\alpha, \lambda}}) \mid \lambda\in[0, 1] \}.
    \end{equation*}
\end{prop}
\begin{sketch}
    Even though $D_\alpha$ is not jointly convex, we can write it as a monotone function of an $f$-divergence, which is jointly convex function and lets us utilize results from multi-objective convex optimization.
\end{sketch}

\section{COMPUTING THE FRONTIERS}
We will now discuss how to compute the divergences when we have access to the distributions.
We discuss strategies for how to do this in practice in \S6.

\subsection{Discrete Measures}\label{subsec:discrete}
\begin{figure*}
\centering
\includegraphics[width=.9\textwidth]{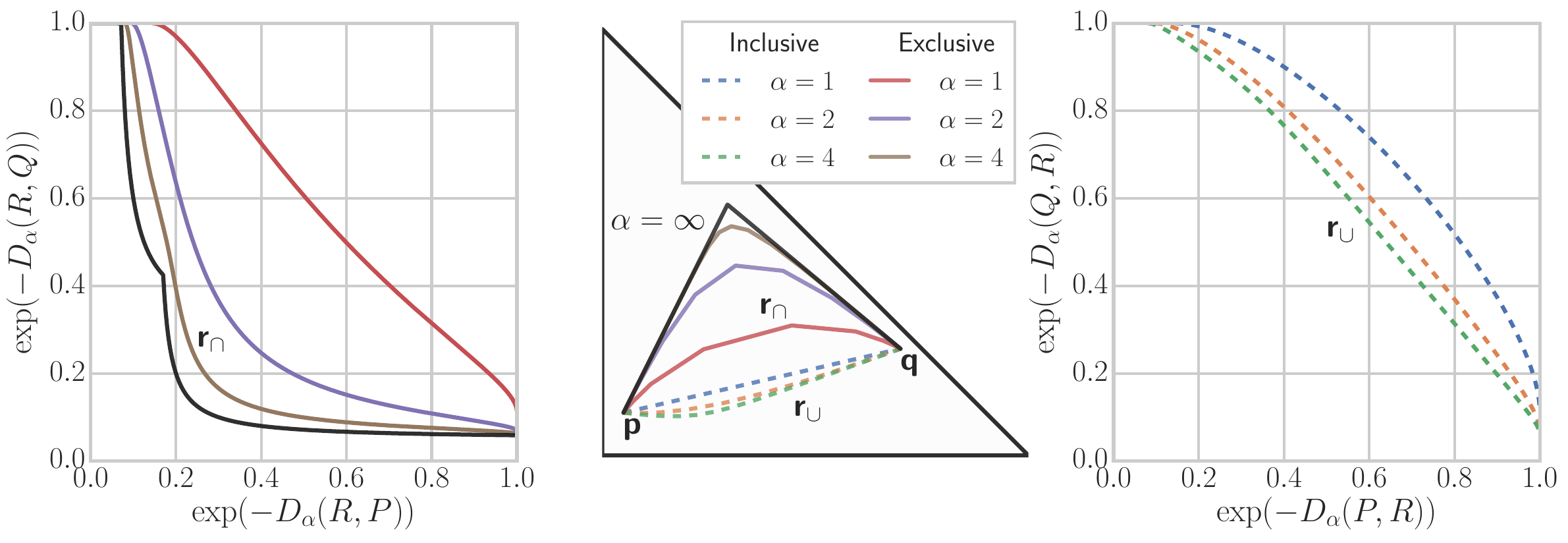}
\caption{Illustration of the algorithm computing the discrete frontiers. In the middle panel we show two measures $\pd$ and $\qd$ on the probability simplex, together with the barycentric paths $\gamma(\lambda)$ between them for various $\alpha$. These paths in turn generate the inclusive (left) and exclusive (right) frontiers. The limiting $\alpha=\infty$ exclusive case coincides with the precision-recall curve from \cite{sajjadi2018assessing} (c.f.~\S\ref{sec:connections}). \label{fig:simplexwalk}}
\end{figure*}

When the distributions take on one of $n$ values, the solution is obtain by simply replacing the integrals with sum in \Cref{prop:discrete-frontiers}.
Hence, if we discretize $\lambda$ over a grid of size $k$, we will have a total complexity of $O(nk)$.
Furthermore, this case has a very nice geometrical interpretation associated with it.
Namely, in this case we can represent the distributions as vectors in the simplex $\Delta = \{ \bmu\in[0,1]^n \mid \bm{1}^\top\bmu = 1 \}$, and use  $\pd\in\Delta$ for $\Pd$ and $\qd\in\Delta$ for $\Qd$.
Then, conceptually, to compute the frontier we walk along the path $\Rd_{\alpha,\gamma}$ from $\pd$ to $\qd$, and at each point we compute the distances to $\bp$ and $\bq$ as measured by $D_\alpha$.
We illustrate this in \Cref{fig:simplexwalk}.

\subsection{Integration}\label{subsec:logratio}

The frontiers can be also written as integrals of functions of the density ratio $p(x)/q(x)$ over the measures $\Pd$ and $\Qd$, which has practical implications, discussed in \Cref{sec:experiments}.
Specifically, we have the following result.

\begin{prop}\label{prop:logratio}
    Define for any $\beta,\lambda$ the functions $u_{\gamma}^\beta(t)=(\gamma + (1-\gamma)t^{\beta})^{(1-\beta)/\beta}$ and $v_{\gamma}^{\beta}(t)=(\gamma + (1-\gamma)t^{\beta})^{1/\beta}$.
    The exclusive frontier $\Ffwd_\alpha(\Pd, \Qd)$ equals
    \begin{align*}
    \{ (&\frac{1}{\alpha-1}\log \mathbb{E}_P[u_{\lambda}^{1-\alpha}(\frac{p(x)}{q(x)})] -
         \frac{\alpha}{\alpha-1} \log \mathbb{E}_P[v_{\lambda}^{1-\alpha}(\frac{p(x)}{q(x)})], \\
       &\frac{1}{\alpha-1} \log\mathbb{E}_Q[u_{1-\lambda}^{1-\alpha}(\frac{q(x)}{p(x)})] - \frac{\alpha}{\alpha-1} \log \mathbb{E}_Q[v_{1-\lambda}^{1-\alpha}(\frac{q(x)}{p(x)})]
    \}.
    \end{align*}
    The inclusive frontier is obtained by changing $u_\gamma^{1-\alpha}$ to $u_\gamma^\alpha$, $v_\gamma^{1-\alpha}$ to $v_\gamma^\alpha$, and $-\alpha/(\alpha-1)$ to $1$.
\end{prop}

Note that the na\"ive plug-in estimator would result in a biased estimate due to the fact that the expectation is inside the logarithm.
Similar problems also arise when doing variational inference with R\'enyi divergences, as discussed in \citet{li2016renyi}, who analyze the bias and the behaviour as a function of the sample size.

\subsection{Exponential families}\label{subsec:continuous}

The computation of the integrals above can be very challenging even if we know the densities due to the possible high dimensionality of the ambient space.
Fortunately, there exists a class of distributions that includes many commonly used distributions called the exponential family, whose frontiers for $\alpha=1$ (i.e., the KL divergence) can be efficiently computed.
This not only includes many popular continuous distributions such as the normal and the exponential, but also many discrete distributions, e.g., tractable Markov Random Fields~\citep{wainwright2008graphical} which are common models in vision and natural language processing. 

\begin{defn}[{{Exponential families \cite[\S 3.2]{wainwright2008graphical}}}] The exponential family over a domain $\mathcal{X}$ for a sufficient statistic $\nu\colon\mathcal{X}\to\mathbb{R}^m$ is the set of all distributions of the form
\begin{equation}\label{eqn:exp-family}
	P(x \mid \btheta) = \exp (\btheta^\top\nu(x)-A(\btheta)),
\end{equation}
where $\btheta$ is the parameter vector, and $A(\btheta)$ is the log-partition function normalizing the distribution.
\end{defn}
Importantly, the KL divergence between two distributions in the exponential family with parameters $\btheta$ and $\btheta'$ can be computed in closed form as the following Bregman divergence \cite[\S 5.2.2]{wainwright2008graphical}
\[
\Dkl{P(\cdot \mid \btheta)}{P(\cdot \mid \btheta')} = A(\btheta')-A(\btheta)-\nabla A(\btheta)^\top(\btheta'-\btheta),
\]
which we shall denote as $\Dkl{\btheta}{\btheta'}$.
We can now show how to compute the frontier.
\begin{prop}\label{prop:exp-families}
Let $\mathcal{M}$ be an exponential family with log-partition function $A$. Let $\Pd$ and $\Qd$ be elements in $\mathcal{M}$ with parameters $\btheta_\Pd$ and $\btheta_\Qd$. Then,
\begin{itemize}
\item Inclusive: If we define $\gamma(\lambda)=(\nabla A)^{-1}(\lambda \nabla A(\btheta_\Pd) + (1-\lambda)\nabla A(\btheta_\Qd))$, then $\Frev_1(\Pd, \Qd \mid \mathcal{M})$ is equal to
\begin{align*}
    \{ (\Dkl{\btheta_\Pd}{\gamma(\lambda)}, \Dkl{\btheta_\Qd}{\gamma(\lambda)}) \mid \lambda\in[0,1] \}.
\end{align*}
\item Exclusive: If we define $\gamma(\lambda)=\lambda \btheta_\Pd + (1-\lambda)\btheta_\Qd$, then $\Ffwd_1(\Pd, \Qd \mid \mathcal{M})$ is equal to
\begin{align*}
    \{ (\Dkl{\gamma(\lambda)}{\btheta_\Pd}, \Dkl{\gamma(\lambda)}{\btheta_\Qd)} &\mid \lambda\in[0,1] \}.
\end{align*}
Furthermore, the frontier will not change if we enlarge $\mathcal{M}$.
\end{itemize}
\end{prop}
\begin{sketch}
    As the KL divergence is convex in the parameter $\btheta'$ of the second distribution, for the inclusive case we can only consider the scalarized objective, which has the claimed closed form solution. In the exclusive case, we use the Bregman divergence generated by the convex conjugate $A^*$, which effectively swaps the arguments, and the argument is the same.
\end{sketch}
 \section{CONNECTIONS TO EXISTING WORK}\label{sec:connections}

\begin{figure}[t]
\centering
\scalebox{0.8}{\centering
\begin{tikzpicture}[scale=2.5]
  \draw[thick, fill=gray!5] (0,0) node[anchor=north]{}
  -- (2*1,0) node[anchor=north]{}
  -- (2*.5,2*.866) node[anchor=south]{}
  -- cycle;

  \node[circle,fill,anchor=north,scale=.4,label=below:$\br$]  (ra) at (.8, .3) {};
  \node[circle,fill,anchor=north,scale=.4,label=right:$\bp$]  (pa) at (.9+.2, .4+.2)  {};
  \node[circle,fill,anchor=north,scale=.4,label=right:$\bp'$] (pb) at (.9+.4, .4+.4)  {};
  \node[circle,fill,anchor=north,scale=.4,label=right:{$\partial_\Delta(\br, \bp)$}]  (pc) at (.9+.55, .4+.55)  {};
  \node[circle,fill,anchor=north,scale=.4,label=left:$\bq$]   (qa) at (.8 + .0, .3 + .3) {};
  \node[circle,fill,anchor=north,scale=.4,label=left:$\bq'$] (qb) at (.8 + .0, .3 + .55) {};
  \node[circle,fill,anchor=north,scale=.4,label=left:{$\partial_\Delta(\br, \bq)$}]   (qc) at (.8 + .0, .3 + 1.1) {};
  \draw (ra) -- (pa) -- (pb);
  \draw (ra) -- (qa) -- (qb);
  \draw[dashed] (pb) -- (pc);
  \draw[dashed] (qb) -- (qc);
  \draw[color=red,thick]  (2*0.55,  2*0.3) -- (2*0.47058824,  2*0.35294118) -- (2*0.44137931,  2*0.33103448) -- (2*0.4,  2*0.3);
\end{tikzpicture}

 }
\caption{The points used in the definition of \PRD~\citep{sajjadi2018assessing}. For fixed $\pd, \qd$ and $\rd$, the points $\pd'$ and $\qd'$ must be lie on the rays $\br\to\bq$ and $\br\to\bp$ respectively. The optimal precision and recall are obtained by taking $\bq'$ and $\br'$ to lie on the boundary. To compute the frontier we have to consider only those $\rd$ on the geodesic between $\bq$ and $\bp$, shown as the red curve.}
\label{fig:prd_special_case}
\end{figure}
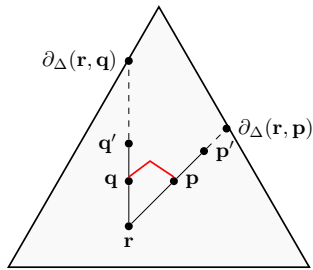

Having introduced and showed how to compute the divergence frontiers, we will now present several existing techniques, and show how they relate to our approach.

\begin{figure}[t]
\centering
\includegraphics[width=.33\textwidth]{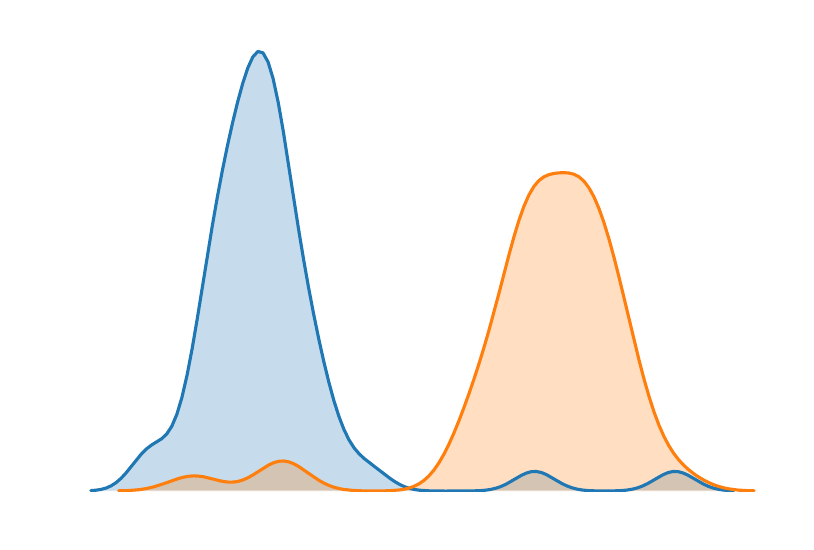}
\caption{
A case where the metric defined by~\cite{kynkaanniemi2019improved} would result in essentially perfect precision and perfect recall. Arguably, these distributions are very different. \label{fig:pr_fail}\vspace{-9pt}}
\end{figure}

Rather than computing trade-off curves, \citet{kynkaanniemi2019improved} focus only on $\Pd(\supp(\Qd))$ and $\Qd(\supp(\Pd))$, and estimate the supports using a union of $k$-nearest neighbourhood balls.
This is indeed a special case of our framework, as $\lim_{\alpha\to 0}\Dalpha{\Pd}{\Qd}=-\log\Qd(\supp(\Pd))$ \cite[Thm.\ 4]{erven2014renyi}.
One drawback of this approach is that all regions where $\Pd$ and $\Qd$ place any mass are considered equal (see Fig.~\ref{fig:pr_fail}).

\looseness=-2 We will now show that the approach of \citep{sajjadi2018assessing} corresponds to the case where $\alpha\rightarrow \infty$. In particular, \citet{sajjadi2018assessing} write both $\Pd$ and $\Qd$
as mixtures with a shared
component that should capture the space that they both assign high
likelihood to, and which can be used to formalize the notions of
precision and recall for distributions.
\begin{defn}[{{\cite[Def.\ 1]{sajjadi2018assessing}}}\label{dfn:prd}] For
$\pi,\rho\in(0,1]$, the probability distribution $\Qd$ has
precision $\pi$ at recall $\rho$ w.r.t.\ $\Pd$ if there exist
distributions $\Rd, \Pd',\Qd'$ such that
\begin{equation}\label{eqn:pr} \Pd = \rho \Rd + (1-\rho)\Pd',
\textrm{ and } \Qd = \pi \Rd + (1-\pi)\Qd'.
\end{equation} The union of $\{(0,0)\}$ and all realizable pairs
$(\pi, \rho)$ will be denoted by $\textsc{PRD}(\Pd, \Qd)$.
\end{defn}

Even though the divergence frontiers introduced in this work might seem unrelated to this formalization, there is a clear connection between them, which we now establish.
As \cite{sajjadi2018assessing} target discrete measures, let us treat the distributions as vectors in the probability simplex
$\Delta$ and use 
$\pd\in\Delta$ for $\Pd$ and $\qd\in\Delta$ for $\Qd$. We need to consider three additional distributions to compute $\PRD(\bp, \bq)$: $\rd$, and the per-distribution mixtures $\pd'$ and $\qd'$.
These distributions are arranged as shown in \Cref{fig:prd_special_case}.
Because $\br, \bp$ and $\bp'$ are co-linear and $\bp=\rho\br + (1-\rho)\bp'$, we have that the recall obtained for this configuration is $\|\bp-\bp'\|/\|\br-\bp'\|$.
Similarly, the precision $\pi$ can be easily seen to be equal to  $\|\bq-\bq'\|/\|\br-\bq'\|$.
Most importantly, we can only
increase both $\rho$ and $\pi$ if we move $\bp'$ and $\bq'$ along the rays $\br\to\bp$ and $\br\to\bq$, respectively. 
Specifically, the maximal recall $\rho^*$ and precision $\pi^*$ for this fixed $\br$ are
obtained when $\bp'$ and $\bq'$ are as far as possible from $\br$,
i.e., when they lie on the boundary $\partial\Delta$.  To formalize
this, let us denote for any $\mathbf{a},\mathbf{b}$ in $\Delta$ by
$\partial_\Delta(\mathbf{a}, \mathbf{b})$ the point along the ray
$\mathbf{a}\to \mathbf{b}$ that intersects the boundary of $\Delta$.  Then, the maximal
$\pi$ and $\rho$ are achieved for $\bp'=\partial_\Delta(\br, \bp)$
and $\bq'=\partial_\Delta(\br, \bq)$.
Perhaps surprisingly, these best achievable precision and recall have been already studied in geometry and have very remarkable properties, as they give rise to a weak metric.

\begin{defn}[{{\cite[Def.~2.1]{papadopoulos2013funk}}}] The Funk weak
metric $F_\Delta\colon\Delta^2\to[0,\infty)$ on $\Delta$ is defined by
\begin{align*} F_\Delta(\bp, \bp) &= 0, \textrm{and } \\
F_\Delta(\bp, \bq) &= \log ( \| \bp - \partial_\Delta(\bp, \bq) \| /
\|\bq - \partial_\Delta(\bp, \bq)\|).
\end{align*}
\end{defn}
Furthermore, we have that the Funk metric coincides with a limiting R\'enyi divergence.
\begin{prop}[{{\cite[Ex.\ 4.1]{papadopoulos2014from},\cite[Thm.~6]{erven2014renyi}}}]
    For any $\bp,\bq$ in the probability simplex $\Delta$, we have that
\[     F_\Delta(\bp, \bq)=\lim_{\alpha\to\infty} \Dalpha{\pd}{\qd} = \log \max_{i=1}^n p_i/q_i
\]
\end{prop}
This immediately implies the following connection between the set of maximal points in $\PRD(\Pd, \Qd)$, which we shall denote by $\overline{\PRD}(\Pd, \Qd)$ and $\mathcal{F}^\cap_\infty(\pd, \qd)$.
In other words, the maximal points in \PRD~coincide with one of the exclusive frontiers we have introduced.
\begin{prop}\label{prop:prd-connection} For any distributions $\Pd,\Qd$ on $\{1, 2, \ldots, n\}$
it holds that
    \[ \overline{\PRD}(\Pd, \Qd) = \{ (e^{-\pi},
e^{-\rho}) \mid (\rho, \pi)\in \mathcal{F}_\infty^\cap(\Pd, \Qd) \}.
    \]
\end{prop}

Furthermore, the fact that $D_\infty$ is a weak metric implies that, in contrast to the $\alpha<\infty$ case, the triangle inequality holds \cite[Thm.\ 7.1]{papadopoulos2013funk}. 
As a result, we can make an even stronger claim --- the path taken by the distributions $\rd$ that generate the frontier is the shortest in the corresponding geometry.

\begin{prop}\label{prop:infinity-geodesic}
	Let us define the curve $\gamma(\lambda)\colon[\min_i \frac{q_i}{p_i}, \max_i\frac{q_i}{p_i}]\to\Delta$ as \[[\gamma(\lambda)]_i \propto \min\{p_i, q_i/\lambda \} . \]
\begin{align*}
   \textrm{Then, }\, \mathcal{F}_\infty^\cap(\pd, \qd)= \{ &(\Dinfty{\gamma(\lambda)}{\bp}, \Dinfty{\gamma(\lambda)}{\bq}) \\ & \mid \lambda\in[\min_i \frac{q_i}{p_i}, \max_i\frac{q_i}{p_i}] \},
\end{align*}
and, moreover, $\gamma(\lambda)$ is geodesic, i.e., it evaluates at the endpoints to $\bp$ and $\bq$, and for any $\lambda$
	\[
	  F_\Delta(\bp, \bq) = F_\Delta(\bp, \gamma(\lambda)) +  F_\Delta(\gamma(\lambda), \bq).
	\]
\end{prop}

\citet{simon2019revisiting} extend this approach to continuous models by showing that that \PRD~ can be computed by thresholding the density ratio $p(x)/q(x)$, which they approximate using binary classification.
In \Cref{subsec:logratio} we have extended this result to arbitrary divergences.

Finally, we note that the idea of precision and recall for generative models
also appeared in~\citet{lucic2018gans} and was used for quantitatively
evaluating generative adversarial networks and variational
autoencoders, by considering a synthetic data set for which the data
manifold is known and the distance from each sample to the manifold
could be computed.

\section{PRACTICAL APPROACHES AND CONSIDERATIONS}\label{sec:experiments}

\begin{figure*}[t]
    \centering
	\begin{subfigure}[b]{.31\textwidth}
	    \includegraphics[width=\textwidth]{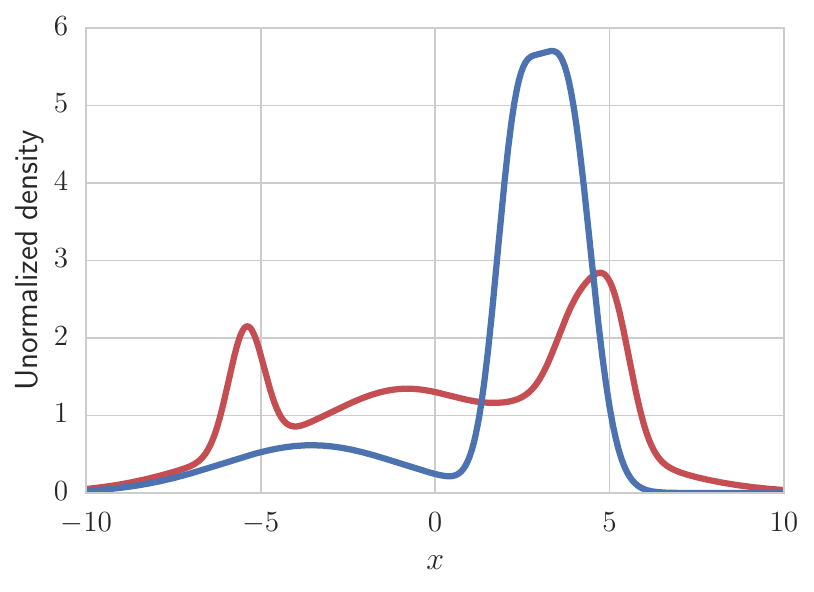}

		\caption{The distributions we quantize.}
	\end{subfigure} 
	\begin{subfigure}[b]{.31\textwidth}
	    \includegraphics[width=\textwidth]{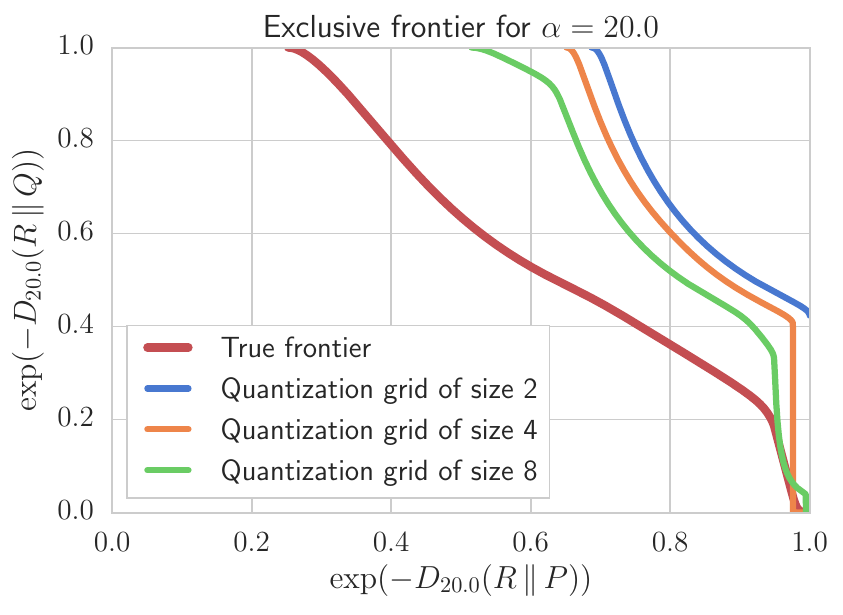}
	    \caption{Biases from coarse quantization.}
	\end{subfigure} 
	\begin{subfigure}[b]{.31\textwidth}
    		\includegraphics[width=\textwidth]{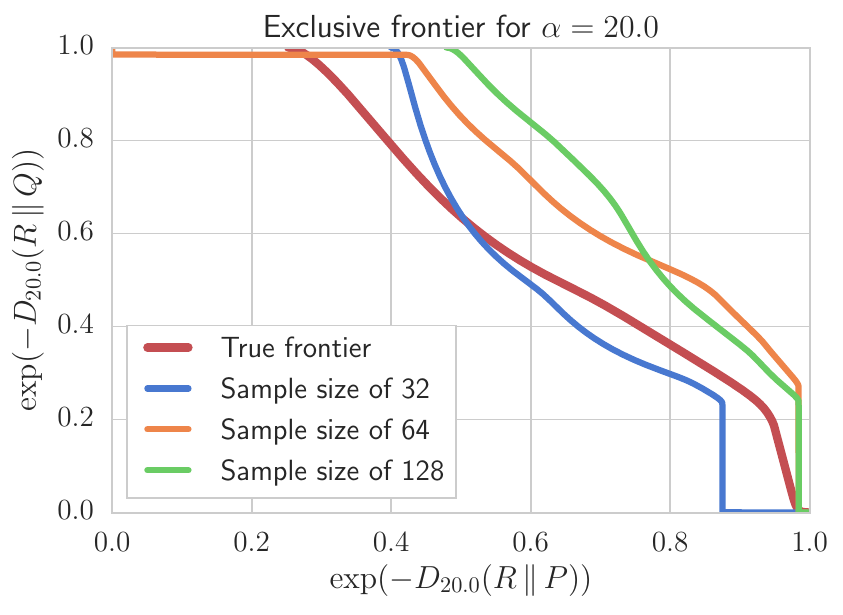}
	    \caption{Noise from small samples.}
	\end{subfigure} 
	\caption{The systemic biases when we compute the frontiers after quantizing the distributions in (a). In (b) we see that using too few buckets can result in overly optimistic results. In (c) we show that if an insufficient number of samples is used, the curves might fluctuate and look  pessimistic.}\label{fig:bias-discrete}
\end{figure*}
\begin{figure*}[t]
    \centering
	\begin{subfigure}[b]{.31\textwidth}
	    \includegraphics[width=\textwidth]{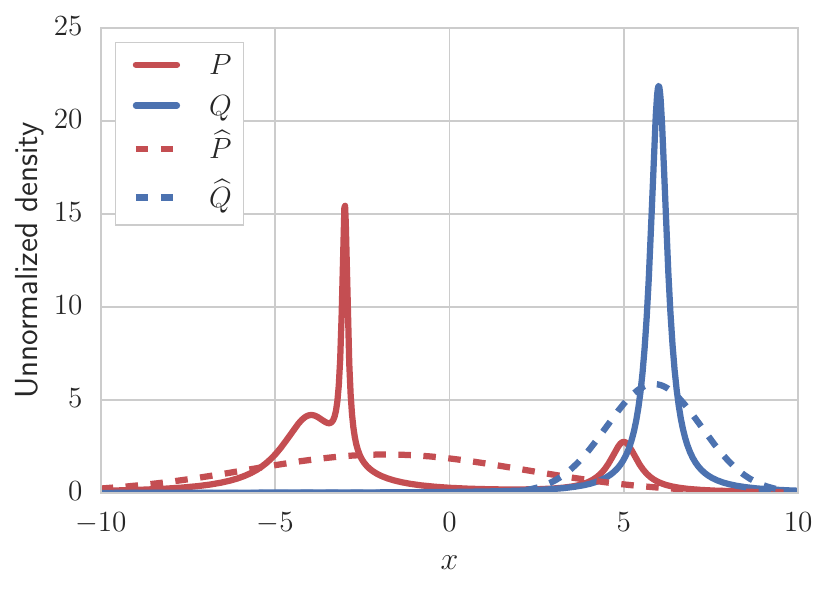}
		\caption{The distributions we compare and their Gaussian approximations.}
	\end{subfigure} 
	\begin{subfigure}[b]{.31\textwidth}
	    \includegraphics[width=\textwidth]{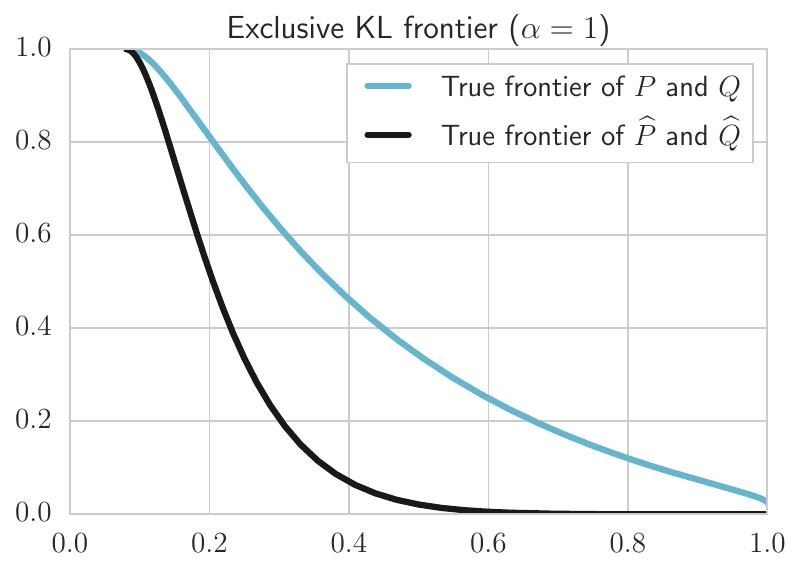}
	    \caption{Estimating the exclusive frontier using Gaussian approximations.}
	\end{subfigure} 
	\begin{subfigure}[b]{.31\textwidth}
    		\includegraphics[width=\textwidth]{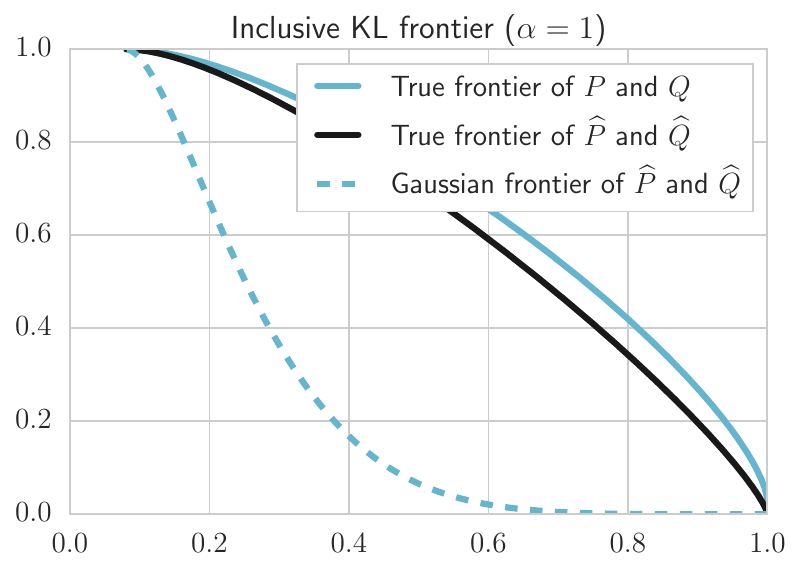}
	\caption{Estimating the inclusive frontier using Gaussian approximations.}
	\end{subfigure} 
	\caption{Estimating the frontier by approximating the distributions with Gaussians. In (b) we see that the exclusive frontier can look more pessimistic than the truth. Panel (c) shows that using distributions $R$ that are further restricted to also be Gaussian can make matters worse when computing the inclusive frontier. Note that in (b) there is no such curve as it agrees with the true frontier (black line) due to the last claim in \Cref{prop:exp-families}.}\label{fig:bias-gaussian}
\end{figure*}

In practice, when we are tasked with the problem of evaluating a model, we typically only have access to samples from $\Pd$ and $\Qd$, and optionally also the density of $\Qd$.
There are many approaches one can undertake when applying the methods developed in this paper to generate precision-recall curves.
In what follows we discuss several of these, and highlight their benefits, but also some of their drawbacks.
We would like to point out that in the case of image synthesis, the comparison is typically not done in the original high-dimensional image space, but (as done in \cite{sajjadi2018assessing,kynkaanniemi2019improved,heusel2017gans}) in feature spaces where distances are expected to correlate more strongly with perceptual difference.

\myparagraph{Quantization}
One strategy would be to discretize the data, as done in \citep{sajjadi2018assessing}, and then apply the methods from \Cref{subsec:discrete}.
Even though estimating the divergences between categorical distributions is simple and in the limit converges to the continuous $\alpha$ divergence \cite[Thm.\ 2]{erven2014renyi}, there may be several issues with this approach.
For example, this approach will inherently introduce a positively bias for any discretization and any $\alpha$.
Namely, the generated curves will always look better than the truth, which we formalize below.

\begin{prop}\label{prop:quantization-bias}
Let $\Pd$ and $\Qd$ be distributions that have been quantized into the discrete distributions $\hat{\Pd}$ and $\hat{\Qd}$ respectively. Then, it holds that
\[
    \mathcal{R}^\cap_\alpha(\hat{\Pd}, \hat{\Qd}) \subseteq
    \mathcal{R}^\cap_\alpha(\Pd, \Qd)
    \textrm{ and }
    \mathcal{R}^\cup_\alpha(\hat{\Pd}, \hat{\Qd}) \subseteq
    \mathcal{R}^\cup_\alpha(\Pd, \Qd).
\]
\end{prop}
Moreover, if we do not have enough samples to estimate the fraction of points that fall in each partition, we might see very strong fluctuations of the curves.
This is due to the fact that the divergences penalize heavily situations when one distributions puts zero mass on the support of the other and the distributions might appear more distant than the truth.
We illustrate both of these situations on two one dimensional distributions in \Cref{fig:bias-discrete}.
We can see that discretization indeed has a positive bias, and that small sample sizes can result in overly pessimistic curves.
Furthermore, in high-dimensions the result will also depend on the quality of the clustering, which  in general is NP-hard and has additional hyperparameters that can be hard to tune.

\myparagraph{Exponential families} Alternatively, one can estimate $\Pd$ and $\Qd$ from samples using maximum likelihood over some exponential family $\mathcal{M}$, and then apply the methods from \Cref{subsec:continuous}, which result in analytical frontiers.
While this might seem simplistic, fitting multivariate Gaussians has been shown to work well for evaluating generative models using the FID score \citep{heusel2017gans}.
Even though projecting $\Pd$ and $\Qd$ onto some exponential family might suggest that it will always make them closer and thus result in a positive bias, this it not necessarily always the case.
We show in \Cref{fig:bias-gaussian} a setting where the opposite happens.
What we can formally show, however, is that the inclusive frontier will have a positive bias when the distribution $\Rd$ we optimize over is restricted to $\mathcal{M}$.

\begin{prop}\label{prop:bias-gaussian}
Let $\Pd$ and $\Qd$ be distributions with maximum likelihood estimates $\hat{P}$ and $\hat{Q}$ belonging to some exponential family $\mathcal{M}$ Then, it holds that
\[
    \mathcal{R}_1^\cup(\Pd, \Qd \mid \mathcal{M}) \subseteq \mathcal{R}_1^\cup(\hat{P}, \hat{Q} \mid \mathcal{M}).
\]
\end{prop}
\begin{sketch}
To show this result we rely on the fact that maximum likelihood estimation is equivalent to (reverse-)projection under the KL divergence, and the fact the KL divergence satisfies a generalized Pythagorean inequality \citep{csiszar2003information}.
\end{sketch}
\vspace{-3mm}
\myparagraph{Density ratio estimation} Similarly to \citep{simon2019revisiting}, one can first estimate the log ratio $p(x)/q(x)$ by fitting a binary classifier \citep{sugiyama2012density}, and then approximate the terms in \Cref{subsec:logratio} using Monte Carlo.
One can also tune the loss function to match the integrands, as suggested by \citet{menon2016linking}.
However, precisely estimating the density ratio is challenging, and large sample sizes might be needed as the estimator is biased.

\myparagraph{Directly estimating $F_1^\cup$} The inclusive frontier for $\alpha=1$ is valid even when we use empirical distributions for $\Pd$ and $\Qd$ without fitting any models.
In this case, it can be easily seen that if we optimize $R_{1,\lambda}^\cup$ over some family $\mathcal{M}$, that this is equivalent to maximum likelihood estimation where the samples come from the mixture $\lambda \Pd + (1-\lambda) \Qd$.
Hence, if we employ flexible density estimators $\mathcal{M}$, one strategy would be to (i) first fit a model on the weighted dataset, and then (ii) evaluate the likelihoods when the data is generated under $P$ and $Q$ on a separate test set.

 \section{CONCLUSIONS}

We developed a framework for comparing distributions via the Pareto frontiers of information divergences, and fully characterized them using efficient computational algorithms for a large family of distributions.
We recovered previous approaches as special cases, and thus provided a novel perspective on them and their algorithms.
Furthermore, we believe that we have also opened many interesting research questions related to classical approximate inference methods --- can we use different divergences or extend the algorithms to even richer model families, and how to identify the correct approach for approximating the frontiers when we only have access to samples.

\section*{Acknowledgements}
We would like to thank Nikita Zhivotovskii for his feedback on the manuscript. We are grateful for the general support and discussions from other members of Google Brain team in Zurich.

\bibliographystyle{apalike}

\newpage
\onecolumn

\appendix
\section{Proofs}
\begin{proof}[Proof of \Cref{prop:prd-connection}]
Even though this result follows clearly from the discussion just above the claim, we provide it for completeness.
Namely, let $(\pi, \rho)\in\overline{\PRD}$ be generated for some $\bp, \bq, \br$.
Based on the argument below \Cref{dfn:prd} it follows that it must be equal to $(\pi,\rho)=(e^{-F_\Delta(\rd, \qd)}, e^{-F_\Delta(\rd, \qd)})$.
Then,  the pair $(\pi,\rho)$ is maximal in $\PRD$ iff $(F_\Delta(\rd, \pd), F_\Delta(\rd, \qd))$ is minimal in $\mathcal{R}^\cap_\infty(\Pd, \Qd)$, i.e., iff $(F_\Delta(\rd, \pd), F_\Delta(\rd, \qd))\in\mathcal{F}^\cap_\infty(\Pd, \Qd)$.
\end{proof}

\begin{proof}[{{Proof of \Cref{prop:infinity-geodesic}}}]
If we also include the normalizer of $\gamma(\lambda)$, we have that
\[
[\gamma(\lambda)]_i = \min\{p_i, q_i/\lambda \}/\beta(\lambda), \textrm{ where } \beta(\lambda)=\sum_{i=1}^n\min\{p_i, q_i/\lambda \}.
\]
The end-point condition is easy to check, namely
\begin{align*}
[\gamma\min_j\{q_j/p_j\})]_i &= \min\{p_i, \frac{q_i}{\min_j\{q_j/p_j\}} \}/\beta(\lambda) = p_i/\beta(\lambda)=p_i, \textrm{ and} \\
[\gamma(\min_j\{q_j/p_j\})]_i &= \min\{p_i, \frac{q_i}{\max_j\{q_j/p_j\}} \}/\beta(\lambda) = q_i/\beta(\lambda)=q_i.
\end{align*}
Let us now show that $\log\beta(\lambda)=-F_\Delta(\gamma(\lambda), \bq)$.
The right hand side can be re-written as
\begin{align*}
    F_\Delta(\gamma(\lambda), \bp)
    = \log \max_i \frac{\min\{p_i, q_i/\lambda \}/\beta(\lambda)}{p_i}
    = - \log \beta(\lambda) + \log \max_i \min\{1, \frac{q_i}{p_i\lambda} \}.
\end{align*}
Note that the term inside the log is not one only if $q_i/p_i<\lambda$ for all $i$, which can happen only if $\lambda>\max_i\frac{q_i}{p_i}$, which is outside the domain of $\gamma$. Similarly,
\begin{align*}
    F_\Delta(\gamma(\lambda), \bq)
    &= \log \max_i \frac{\min\{p_i, q_i/\lambda \}/\beta(\lambda)}{q_i} \\
    &= - \log \beta(\lambda) + \log \max_i \min\{\frac{p_i}{q_i}, 1/\lambda \} \\
    &= - \log \beta(\lambda)\lambda + \log \max_i \min\{\frac{\lambda p_i}{q_i}, 1\}.
\end{align*}
The claim follows because $\alpha(\lambda) = \lambda\beta(\gamma)$, and by noting that the maximum inside the logarithm is strictly less than one only if for all $i$ it holds that $\lambda < \frac{q_i}{p_i}$, which is outside the domain of $\gamma$.

Finally, let us show the geodesity of the curve.
\begin{align*}
F_\Delta(\bp, \bmu^*(\lambda)) + F_\Delta(\bmu^*(\lambda), \bq)
&=
\log\max_i\frac{p_i}{\min\{ p_i, q_i/\lambda \}/\beta(\lambda)} +
\log\max_i\frac{\min\{ p_i, q_i/\lambda \}/\beta(\lambda)}{q_i} \\
&= \max_i \log \max \{\log \frac{\lambda p_i}{q_i}, 1 \} + \max_i \log \min \{\frac{p_i}{q_i}, \frac{1}{\lambda}\}
\end{align*}
\begin{itemize}
\item \emph{Case (i): $\lambda\geq \max_i \frac{q_i}{p_i}$.} Then, $\frac{\lambda p_i}{q_i}\lambda \geq 1$, so that the first term will be equal to $\log \lambda + \log\max_i\frac{p_i}{q_i}$. Similarly, $\lambda^{-1}\leq \frac{p_i}{q_i}$, so that the second term is equal to $-\log\lambda$, and the claimed equality is satisfied.

\item \emph{Case (ii): $\lambda<\max_i \frac{q_i}{p_i}$.} Note that
\begin{align*}
 &\max_i \log \max \{\log \frac{\lambda p_i}{q_i}, 1 \} + \max_i \log \min \{\frac{p_i}{q_i}, \frac{1}{\lambda}\} = \\
 &\max_i \log \lambda \max \{\log \frac{p_i}{q_i}, 1/\lambda \} + \max_i \log \frac{1}{\lambda} \min \{\frac{p_i\lambda }{q_i}, 1\},  
\end{align*}
so that the problem is symmetric if we parametrize with $\lambda'=\lambda^{-1}$ and the argument from above holds.
\end{itemize}
\end{proof}

\begin{proof}[\Cref{prob:exact-solutions}]
We have that
\begin{align*}
\lambda \hatDalpha{\Qd}{\Rd} + (1-\lambda)\hatDalpha{\Pd}{\Rd} &=
\frac{1}{1-\alpha} \int \lambda q(x)^\alpha r(x)^{1-\alpha}dx +
\frac{1}{1-\alpha} \int (1-\lambda) p(x)^\alpha r(x)^{1-\alpha}dx \\ &=
\frac{1}{1-\alpha} \int (\lambda q(x)^\alpha + (1-\lambda) p(x)^\alpha) r(x)^{1-\alpha}dx \\ &=
\frac{1}{1-\alpha} \int ((\lambda q(x)^\alpha + (1-\lambda) p(x)^\alpha)^{1/\alpha})^\alpha r(x)^{1-\alpha}dx \\ &=
\hatDalpha{\Rd_{\alpha,\lambda}^\cap}{\Rd} \int (\lambda q(x)^\alpha + (1-\lambda p(x)^\alpha))^{1/\alpha}dx,
\end{align*}
from which the claim follows as $\hat{D}_\alpha$ is an $f$-divergence and thus minimal and equal to zero only when its arguments agree, and the second term is a constant with respect to $R$.
The other case can be similarly shown by replacing $\alpha$ with $1-\alpha$, namely
\begin{align*}
\lambda \hatDalpha{\Rd}{\Qd} + (1-\lambda)\hatDalpha{\Rd}{\Pd} &=
\frac{1}{1-\alpha} \int \lambda q(x)^{1-\alpha} r(x)^{\alpha}dx +
\frac{1}{1-\alpha} \int (1-\lambda) p(x)^{1-\alpha} r(x)^{\alpha}dx \\ &=
\frac{1}{1-\alpha} \int (\lambda q(x)^{1-\alpha} + (1-\lambda) p(x)^{1-\alpha}) r(x)^{\alpha}dx \\ &=
\frac{1}{1-\alpha} \int ((\lambda q(x)^{1-\alpha} + (1-\lambda) p(x)^{1-\alpha})^{1/(1-\alpha)})^{1-\alpha} r(x)^{\alpha}dx \\ &=
\hatDalpha{\Rd}{\Rd_{\alpha,\lambda}^\cup} \int (\lambda q(x)^{1-\alpha} + (1-\lambda p(x)^{1-\alpha}))^{1/(1-\alpha)}dx.
\end{align*}
\end{proof}

\begin{proof}[{{Proof of \Cref{prop:discrete-frontiers}}}]
\emph{Case (i)} Remember that we want to minimize $R\to\Dalpha{R}{P}$ and $R\to\Dalpha{R}{Q}$.
We want to optimize over the set of all distributions $R$ that have a density so that the integrals are well-defined.
Instead of minimizing the R\'enyi divergences $\frac{1}{\alpha-1}\log\int (r(x)/q(x))^{\alpha-1}r(x)dx$, we can alternatively minimize the $\alpha$-divergences 
$\hat{R}_\alpha$ as they are monotone functions of each other, as already mentioned above \Cref{prob:exact-solutions}.
As the $\alpha$ divergence is an $f$-divergence (see e.g.~\cite[C]{nielsen2009dual}), it follows that it is jointly convex in both arguments.
Hence the Pareto frontier can be computed using the linearly scalarized problem (for a proof see \cite[\S 4.7.3]{boyd2004convex}).
The claim then follows from \Cref{prob:exact-solutions}.

\emph{Case (ii)} This case follows analogously as above as the $f$-divergence is jointly convex, and using the corresponding result from \Cref{prob:exact-solutions}.
\end{proof}

\begin{proof}[{{Proof of \Cref{prop:exp-families}}}]
The proof follows the same argument of \citet[\S 2]{nielsen2007centroids}, the main difference that we also discuss about Pareto optimality, while in the \cite{nielsen2007centroids} the authors only discuss the barycenter problem.
Let us denote for any convex continuously differentiable function $G\colon\R^d\to\R$ by $B_G\colon\R^d\times \R^d\to\R$ the Bregman divergence generated by $G$, i.e.,
\[
    B_G(\mathbf{x}, \mathbf{y}) = F(\mathbf{x}) - F(\mathbf{y}) - \nabla F(\mathbf{y})^\top(\mathbf{x} - \mathbf{y}).
\]
In the inclusive case, we want to minimize the objectives $\btheta_\Rd\to \Dalpha{\btheta_\Pd}{\btheta_\Rd}$ and $\btheta_\Rd\to \Dalpha{\btheta_\Qd}{\btheta_\Rd}$ over $\btheta_\Rd$.
In terms of Bregman divergences, we want to minimize $B_A(\btheta_\Rd, \btheta_\Pd)$ and $B_A(\btheta_\Rd, \btheta_\Qd)$.
Because Bregman divergences are convex in their first argument, as in the proof of \Cref{prop:discrete-frontiers} we can only consider the solutions to the linearly scalarized objective
\[
   \lambda B_A(\btheta_\Rd, \btheta_\Pd) + (1-\lambda) B_A(\btheta_\Rd, \btheta_\Qd),
\]
whose solution is known (see e.g.~\cite{banerjee2005clustering}) to be equal to $\btheta^*_\Rd(\lambda)=\lambda \btheta_\Pd + (1-\alpha)\btheta_\Qd$, which we had to show.
The exclusive case follows from the same argument using the fact that $B_A(\btheta, \btheta') = B_{A^*}(\nabla A(\btheta'), \nabla A(\btheta))$ and that $\nabla A^*=(\nabla A)^{-1}$ \cite[Prop.\ B.2]{wainwright2008graphical}.

The final claim follows from \citep[Lemma 6.6]{van2010data}, which shows that the fact that the optimal $R$ is given by the distribution with density $r(x)\propto p(x)^\lambda q(x)^{1-\lambda}$, which is a member of the exponential family and has a parameter $\lambda \btheta_\Pd + (1-\lambda)\btheta_\Qd$.
\end{proof}

\begin{proof}[\Cref{prop:quantization-bias}]
This follows directly from \citep[Theorem 10]{erven2014renyi} which claims that for any two distributions $\Pd$ and $\Qd$ and any $\alpha$ it holds that $\Dalpha{\Pd}{\Qd} = \sup_{\mathcal{P}}\Dalpha{\Pd_{|\mathcal{P}}}{\Qd_{|\mathcal{P}}}$, where $\mathcal{P}$ is any partition of the $\sigma$-algebra over which the measures are defined. 
\end{proof}

\begin{proof}[\Cref{prop:bias-gaussian}]
The distributions $\hat{\Pd}$ and $\hat{\Qd}$ are maximum likelihood estimators of $\Pd$ and $\Qd$ respectively.
This means that they minimize $\Dkl{\Pd}{\Rd}$ and $\Dkl{\Qd}{\Rd}$ over $\Rd\in\mathcal{M}$ and are thus right information projections onto $\mathcal{M}$ \citep{csiszar2003information}.
Then, as exponential families are log-convex, from \citep[Theorem 1]{csiszar2003information} it follows that for any $R\in\mathcal{M}$ we have that $\Dkl{\Pd}{\Rd} \geq \Dkl{\hat{\Pd}}{\Rd}$ and  $\Dkl{\Qd}{\Rd} \geq \Dkl{\hat{\Qd}}{\Rd}$, which directly implies the result.
\end{proof}

\begin{proof}[\Cref{prop:logratio}]
The results are algebraic manipulations that directly follow from  \Cref{prop:discrete-frontiers} and \Cref{prob:exact-solutions}, and are provided here for completeness.

Let us first compute the terms for the inclusive frontier.
\begin{align*}
(\alpha-1) \Dalpha{\Pd}{\R_{\alpha,\lambda}^\cup} &=
\log \int p(x)^{\alpha}\frac{1}{Z^{1-\alpha}} (\lambda q(x)^{\alpha} + (1 - \lambda) p(x)^\alpha)^{\frac{1-\alpha}{\alpha}} dx \\ &=
\log \int p(x) p(x)^{\alpha-1}\frac{1}{Z^{1-\alpha}} (\lambda q(x)^{\alpha} + (1 - \lambda) p(x)^\alpha)^{\frac{1-\alpha}{\alpha}} dx \\ &=
\log \int p(x) (\lambda (q(x)/p(x))^{\alpha} + (1 - \lambda))^{\frac{1-\alpha}{\alpha}} dx + (\alpha-1)\log Z,
\end{align*}
where
\begin{align*}
  \log Z &= \int (\lambda q(x)^{\alpha} + (1-\lambda) p^\alpha(x))^{1/\alpha}dx \\
    &= \int (\lambda q(x)^\alpha+ (1-\lambda)p^\alpha(x))^{1/\alpha}p(x)^{-\alpha/\alpha} p(x)dx \\
    &= \int (\lambda (q(x)/p(x))^\alpha + (1-\lambda))^{1/\alpha}p(x)dx \\
\end{align*}
which equals the claimed form. The other coordinate of the frontier is obtained by swapping $\Pd$ with $\Qd$ and $\lambda$ with $1-\lambda$.
The equations for the exclusive frontier are obtained by replacing $\alpha$ with $1-\alpha$ on the right hand sides of the above equations.
\end{proof}

\end{document}